\documentclass[conference]{IEEEtran}
\usepackage{lineno,hyperref}
\usepackage{multirow}
\usepackage{graphicx}
\usepackage{amsmath}
\usepackage{amssymb}

\ifCLASSINFOpdf
 
\begin{document}

\title{Breast mass classification in ultrasound based on Kendall's shape manifold}

\author{\IEEEauthorblockN{Micha\l{} Byra\IEEEauthorrefmark{1}\IEEEauthorrefmark{2}\IEEEauthorrefmark{3}, Michael Andre\IEEEauthorrefmark{3}}

\IEEEauthorblockA{\IEEEauthorrefmark{2}Department of Ultrasound, Institute of Fundamental Technological Research, \\Polish Academy of Sciences, Warsaw, Poland}
\IEEEauthorblockA{\IEEEauthorrefmark{3}Department of Radiology, University of California, San Diego, USA}
\IEEEauthorblockA{\IEEEauthorrefmark{1}Corresponding author, e-mail: mbyra@ippt.pan.pl}
}

\maketitle

\begin{abstract}
 
 Morphological features play an important role in breast mass classification in sonography. While benign breast masses tend to have a well-defined ellipsoidal contour, shape of malignant breast masses is commonly ill-defined and highly variable. Various handcrafted morphological features have been developed over the years to assess this phenomenon and help the radiologists differentiate benign and malignant masses. In this paper we propose an automatic approach to morphology analysis, we express shapes of breast masses as points on the Kendall's shape manifold. Next, we use the full Procrustes distance to develop support vector machine classifiers for breast mass differentiation. The usefulness of our method is demonstrated using a dataset of B-mode images collected from 163 breast masses. Our method achieved area under the receiver operating characteristic curve of 0.81. The proposed method can be used to assess shapes of breast masses in ultrasound without any feature engineering. 

\end{abstract}


\IEEEpeerreviewmaketitle

\section{Introduction}

Breast mass shape in 2D ultrasound (US) imaging is an important factor taken into account by radiologists to accurately differentiate malignant and benign breast masses. While benign masses tend to have well-defined ellipsoidal shape, malignant masses usually have variable contour, which includes lobulations and spiculations \cite{rahbar1999benign}. Assessment of breast mass shape is an important part of the Breast Imaging Reporting and Data System (BI-RADS), which was developed by American College of Radiology to standardize the reporting process and diagnosis of breast masses in US. To help the radiologists assess breast masses in US various computer aided diagnosis systems have been proposed \cite{GomezFlores20151125}. Those system commonly aim to computerize shape related information to extract morphological features that would be useful for the classification \cite{alvarenga2012assessing,chen2003breast,chou2001stepwise}. For example, features related to breast mass width-to-depth ratio, circularity and boundary roughness were reported to be efficient for the differentiation of malignant and benign breast masses \cite{GomezFlores20151125}. 

In this work we propose a novel approach to breast mass shape analysis in US. Our approach does not require feature engineering, therefore is more automatic than the previous methods based on the handcrafted morphological features. To assess breast mass shape we utilize Kendall's shape manifold \cite{kendall1984shape}. In Kendall's formalism, shape is related to geometrical properties of a 2D object, which are invariant to rotation, scale and translation. Shape of any object can be represented as a set of landmarks, points on object contour that capture shape variability. This set of landmarks corresponds to a point on the shape manifold. Similarly, shape of a breast mass can be expressed as a point on the Kendall's shape manifold using landmarks. Next, by defining a distance function between points on the manifold it is possible to develop a classification method \cite{jayasumana2013framework,mtibaa2017bayesian}. Shapes of different objects can be compared without engineering features that would accurately describe geometrical properties of those objects. For example, the Procrustes distance function can be used to develop a support vector machine (SVM) classifier that can differentiate 2D shapes specified by landmarks \cite{jayasumana2013framework}. As far as we know, this is the first application of the Kendall's shape theory in US imaging. Image recognition methods based on the Kendall's shape manifold were employed, for example, for plant species classification \cite{jayasumana2013framework,mtibaa2017bayesian}. Here, we use a similar approach to the one proposed in \cite{jayasumana2013framework} to address the problem of breast mass classification. We express contours of breast masses as sets of landmarks and develop SVM classifiers that can directly differentiate malignant and benign masses. 

\section{Materials and Methods}

\subsection{Dataset}

\begin{figure}[t]
	\begin{center}
		\includegraphics[width=1.0\linewidth]{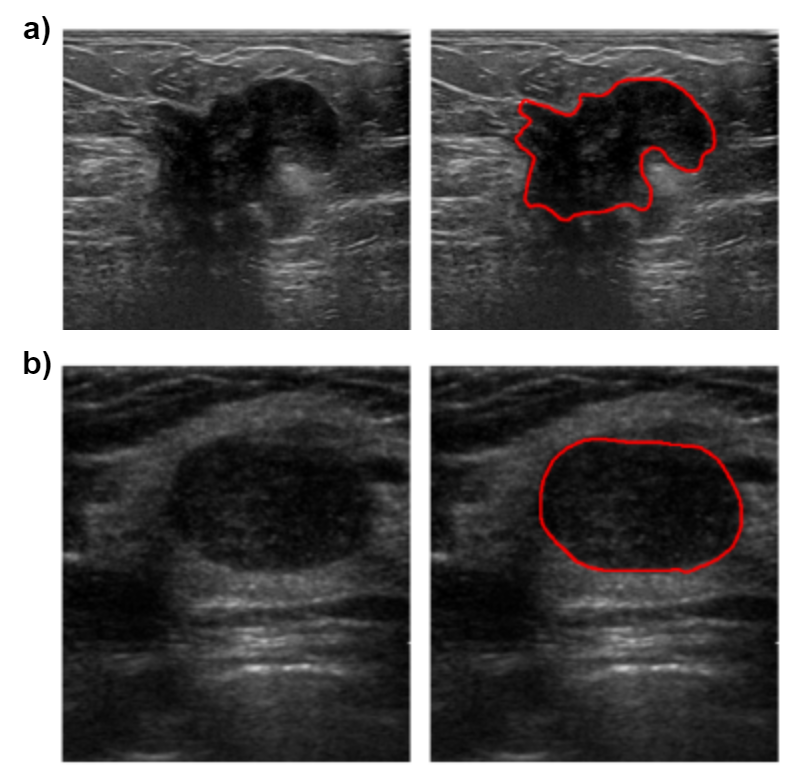}
	\end{center}
	\caption{Ultrasound B-mode images corresponding to a) malignant mass and b) benign mass, and the corresponding manual segmentations indicating breast mass shapes. Contours of malignant masses are commonly more variable than for benign cases.}
	\label{fig:bus1}
\end{figure}

To investigate the usefulness of the Kendall's shape manifold for breast mass classification we used a publicly available dataset of US breast mass images named UDIAT \cite{yap2018automated}. This dataset includes 163 US images collected from 53 malignant and 110 benign breast masses, respectively. The dataset was originally used to develop algorithms for breast mass detection and segmentation \cite{yap2018automated, yap2018breast}. In the case of the UDIAT dataset each B-mode image comes with a region of interest (ROI) indicating breast mass position (Fig. \ref{fig:bus1}). 

\subsection{Shape analysis}

\begin{figure}[t]
	\begin{center}
		\includegraphics[width=1.0\linewidth]{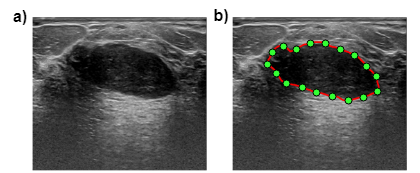}
	\end{center}
	\caption{Ultrasound B-mode image of a benign mass with 18 equidistant landmarks indicating shape.}
	\label{fig:bus2}
\end{figure}

Contour of a 2D shape can be represented by a finite set of landmarks, specific and unique points on contour. Each landmark is coded by a single complex number, with the real and imaginary parts corresponding to axial and lateral landmark coordinates, therefore for $n$ landmarks shape of  an object is initially represented as a $n$-dimensional complex vector \cite{kendall1984shape}. Fig. \ref{fig:bus2} presents a breast mass with landmarks indicating points on contour. In this case, the distances between the consecutive points measured along the contour were the same. Next, to make the landmark shape representation invariant to scaling and translation, the complex vector is processed to have zero mean and unit norm. This shape representation, after the processing, is dubbed pre-shape, and denoted by $\mathbf{z}$. Additionally, to remove rotation different rotated versions of pre-shape are determined and final shape corresponds to the equivalence class of pre-shapes, denoted by $[\mathbf{z}]$ \cite{jayasumana2013framework}.

The similarity of two shapes $[\mathbf{z}_1]$ and $[\mathbf{z}_2]$ can be assessed with the full Procrustes (FP) distance:

\begin{equation}
    d_{FP}([\mathbf{z}_1], [\mathbf{z}_2]) = (1-|\langle \mathbf{z}_1,\mathbf{z}_2\rangle|^2)^\frac{1}{2},
\end{equation}

\noindent where $\mathbf{z}_1$ and $\mathbf{z}_2$ are the corresponding pre-shapes, $\langle \cdot, \cdot \rangle $ and $|\cdot|$ denote the inner product and complex number norm, respectively. The FP distance can be used to define a positive definite kernel function based on the Gaussian radial basis function \cite{jayasumana2013framework}:

\begin{equation}
   K([\mathbf{z}_i], [\mathbf{z}]) = \text{exp}(-d_{FP}^2([\mathbf{z}_1], [\mathbf{z}_2])/2\sigma^2),
\end{equation}

\noindent where $\sigma>0$ is the kernel parameter. Points on the shape manifold can be mapped to elements in a high dimensional Hilbert space. This Hilbert space embedding of shapes can be utilized to develop a binary SVM classifier. Thanks to the kernel trick, the SVM classifier can be expressed in the following dual form \cite{cristianini2000introduction}:

\begin{equation}
    f(\mathbf{[z]}) = \sum_{i=1}^{N} \alpha_i c_i K([\mathbf{z}_i], [\mathbf{z}]),
\end{equation}

\noindent where $N$ is the number of training samples, $\alpha_i$ are the weights corresponding to each sample, $c_i \in \{-1,1\}$ are the corresponding sample classes (benign or malignant in our case) and $K(\cdot,\cdot)$ is the kernel function from eq. 2, which is a valid kernel function for the SVM algorithm \cite{jayasumana2013framework}. The samples with non-zero $\alpha_i$ are denoted support vectors.

\subsection{Statistical analysis}

To evaluate the proposed method we applied leave-one-out cross-validation. The SVM classifiers were developed for breast mass shapes coded with different numbers of landmarks equal to 25, 50, 100 and 200, respectively. The classification performance was assessed using the receiver operating characteristic (ROC) curve and the area under the ROC curve. Accuracy, sensitivity and specificity of each classifier were calculated using the classifier cut-off value corresponding to the point on ROC curve, which was the closest to curve upper left corner, point (0, 1). Differences in AUC values were compared using the DeLong test implemented in R \cite{delong1988comparing,robin17proc}. Calculations, including implementation of the SVM classifier, were done in Matlab (Mathworks, MA).

\section{Results}

\begin{figure}[t]
	\begin{center}
		\includegraphics[width=.8\linewidth]{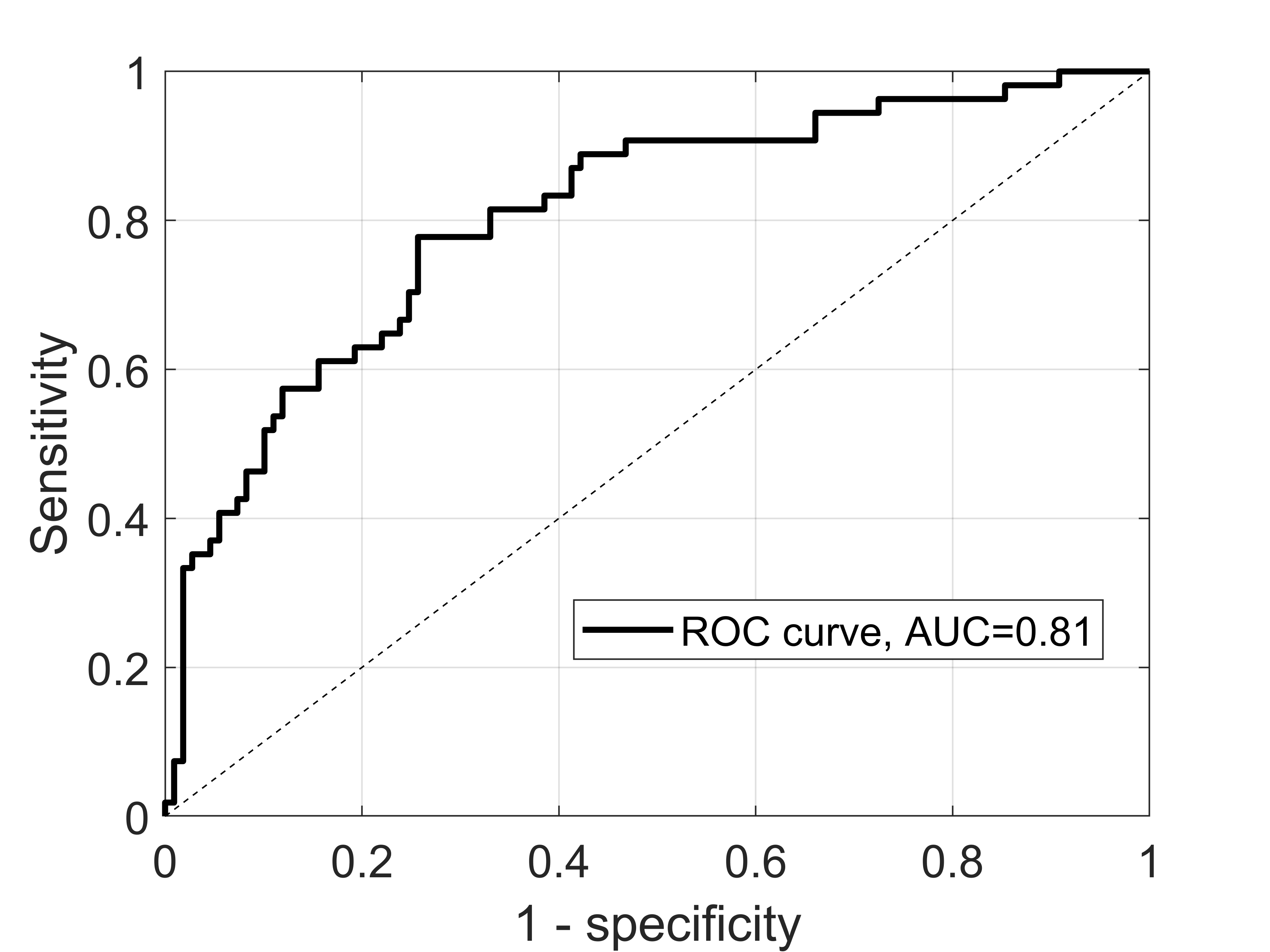}
	\end{center}
	\caption{The receiver operating characteristic (ROC) curve calculated for the proposed method (200 landmarks) utilizing Kendall's shape manifold, with the area under the curve (AUC) equal to 0.81.}
	\label{fig:bus3}
\end{figure}

Table 1 presents the breast mass classification performance results obtained for the proposed method. The SVM classifiers utilizing different number of landmarks achieved AUC values of around 0.81. There were no associated differences in the AUC values calculated for the SVM classifiers ($p$-values$>$0.05). The ROC curve obtained for the method based on 200 landmarks is presented in Fig. \ref{fig:bus3}.  

\begin{table}[t]
\caption{Classification performance of the proposed method based on Kendall's shape manifold obtained for different number of used landmarks. AUC - area under the receiver operating characterstic curve}
	\begin{center}
		\begin{tabular}{|c|c|c|c|c|}
			\hline
			 \# Landmarks & AUC & Accuracy & Sensitivity & Specificity \\
			\hline \hline
			25 & 0.81$\pm$0.03 & 0.81$\pm$0.2 & 0.70$\pm$0.05 & 0.74$\pm$0.04 \\
             \hline
			50 & 0.81$\pm$0.03 & 0.76$\pm$0.1 & 0.74$\pm$0.04 & 0.75$\pm$0.04 \\
             \hline
			100 &  0.81$\pm$0.03 & 0.76$\pm$0.1 & 0.78$\pm$0.04 & 0.74$\pm$0.03 \\
             \hline
            200 &  0.81$\pm$0.03 & 0.76$\pm$0.1 & 0.78$\pm$0.04 & 0.74$\pm$0.04 \\
             \hline
		\end{tabular} 
	\end{center}
	\label{tab:X1}
\end{table}

\section{Discussion}

In this work we presented the feasibility of using Kendall's shape manifold for breast mass classification in US. Our approach, based on the Gaussian kernel with the FP distance and the SVM algorithm, achieved AUC value of 0.81, indicating medium classification performance. Our approach has several advantages. First of all, standard methods usually require expert knowledge to develop better performing morphological features. In comparison, the proposed approach is semi-automatic. In our case the problem of feature engineering is replaced with the task how to efficiently select landmarks and classification method. Currently, deep learning methods are gaining momentum for breast mass classification \cite{antropova2017deep,byra2017combining,byra2018discriminant,byra2019breast,han2017deep,qi2019automated}. These deep learning methods can process US images fully automatically to extract image features related to malignancy and perform classification. While our method require to first manually segment each breast mass, in comparison to deep learning methods our approach does not require large dataset to develop the classifier.  

There are also several issues related to our approach. First, our method, like the standard approaches based on morphological features, still depends on the quality of the initial manual segmentation. Moreover, the interpretability of the proposed machine learning method is lower than in the case of the methods based on expert knowledge. Morphological features related to shape properties, such as circularity or depth-to-width ratio, have clear interpretation. Nevertheless, for a test breast mass US image, our approach can be used to retrieve similar US images from the training set to help understand the classification decision. Another issue is related to the shape representation in Kendall's formalism. This representation is invariant to scale and rotation, but those two properties might be related to breast mass malignancy \cite{GomezFlores20151125}. 

In future we would like to compare the performance of our approach with a classifier developed using morphological handcrafted features. Moreover, in our work we used only one approach to shape analysis based on landmark representation, but it would be interesting to explore the usefulness of approaches based on other representations, for example those related to level sets \cite{cremers2004kernel}.

\section{Conclusions}

In this work we proposed a novel approach to breast mass differentiation in ultrasound. Our method, based on the Kendall's shape manifold, achieved medium classification performance. The proposed approach can be used to assess breast mass shapes without any feature engineering. 

% \section*{Acknowledgement} 

\section*{Conflict of interest statement}

The authors do not have any conflicts of interest. 

\bibliographystyle{IEEEtran}
\bibliography{mybibfile}

\end{document}